\documentclass{article}

\usepackage{microtype}
\usepackage{graphicx}
\usepackage{subfigure}
\usepackage{booktabs} 
\usepackage{amsfonts, amsmath} 
\usepackage{graphicx, subfigure, bm}

\newcommand{\x}{\bm{x}}
\newcommand{\z}{\bm{z}}
\newcommand{\f}{\bm{f}}

\usepackage{hyperref}

\usepackage[accepted]{icml2018}

\icmltitlerunning{Disentangled Sequential Autoencoder}

\begin{document}

\twocolumn[
\icmltitle{Disentangled Sequential Autoencoder}




\icmlsetsymbol{equal}{*}

\begin{icmlauthorlist}
\icmlauthor{Yingzhen Li}{cam}
\icmlauthor{Stephan Mandt}{dr}
\end{icmlauthorlist}

\icmlaffiliation{cam}{University of Cambridge}
\icmlaffiliation{dr}{Disney Research, Los Angeles, CA, USA}

\icmlcorrespondingauthor{Yingzhen Li}{yl494@cam.ac.uk}
\icmlcorrespondingauthor{Stephan Mandt}{stephan.mandt@gmail.com}

\icmlkeywords{Machine Learning, ICML}

\vskip 0.3in
]



\printAffiliationsAndNotice{}  

\begin{abstract}
We present a VAE architecture for encoding and generating high dimensional sequential data, such as video or audio. Our deep generative model learns a latent representation of the data which is split into a static and dynamic part, allowing us to approximately disentangle latent time-dependent features (dynamics) from features which are preserved over time (content). This architecture gives us partial control over generating content and dynamics by conditioning on either one of these sets of features. In our experiments on artificially generated cartoon video clips and voice recordings, we show that we can convert the content of a given sequence into another one by such content swapping. For audio, this allows us to convert a male speaker into a female speaker and vice versa, while for video we can separately manipulate shapes and dynamics.  Furthermore, we give empirical evidence for the hypothesis that stochastic RNNs as latent state models are more efficient at compressing and generating long sequences than deterministic ones, which may be relevant for applications in video compression.
\end{abstract}

\section{Introduction}
\label{sec:intro}
Representation learning remains an outstanding research problem in machine learning and computer vision. Recently there is a rising interest in disentangled representations, in which each component of learned features refers to a semantically meaningful concept. In the example of video sequence modelling, an ideal disentangled representation would be able to separate time-independent concepts (e.g.~the identity of the object in the scene) from dynamical information (e.g.~the time-varying position and the orientation or pose of that object). Such disentangled representations would open new efficient ways of compression and style manipulation, among other applications.

Recent work has investigated disentangled representation learning for images within the framework of variational auto-encoders (VAEs) \citep{kingma2013auto,rezende2014stochastic} and generative adversarial networks (GANs) \citep{goodfellow2014generative}. Some of them, e.g.~the $\beta$-VAE method \citep{higgins2016early}, proposed new objective functions/training techniques that encourage disentanglement. 
%
On the other hand, network architecture designs that directly enforce factored representations have also been explored by e.g.~\citet{siddharth2017learning, bouchacourt2017multi}. 
These two types of approaches are often mixed together, e.g.~the infoGAN approach \citep{chen2016infogan} partitioned the latent space and proposed adding a mutual information regularisation term to the vanilla GAN loss. \citet{mathieu2016disentangling} also partitioned the encoding space into style and content components, and performed adversarial training to encourage the datapoints from the same class to have similar content representations, but diverse style features.

Less research has been conducted for \emph{unsupervised} learning of disentangled representations of sequences. For video sequence modelling, \citet{villegas2017decomposing} and \citet{denton2017unsupervised} utilised different networks to encode the content and dynamics information separately, and trained the auto-encoders with a combination of reconstruction loss and GAN loss. Structured~\citep{johnson2016composing} and Factorised VAEs~\citep{deng2017factorized} used hierarchical priors to learn more interpretable latent variables. \citet{hsu2017unsupervised} designed a structured VAE in the context of speech recognition. Their VAE architecture 
is trained using a combination of the standard variational lower bound and a discriminative regulariser to further encourage disentanglement. More related work is discussed in Section~\ref{sec:related}.

In this paper, we propose a generative model for unsupervised structured sequence modelling, such as video or audio. 
We show that, in contrast to previous approaches, a disentangled representation can be achieved by a careful design of the probabilistic graphical model.
In the proposed architecture, we explicitly use a latent variable to represent \emph{content}, i.e., information that is invariant through the sequence, and a set of latent variables associated to each frame to represent \emph{dynamical} information, such as pose and position. Compared to the mentioned previous models that usually predict future frames conditioned on the observed sequences, we focus on learning the distribution of the video/audio content and dynamics to enable sequence generation without conditioning. Therefore our model can also generalise to unseen sequences, which is confirmed by our experiments. In more detail, our contributions are as follows:
\begin{itemize}
\item {Controlled generation.} Our architecture allows us to approximately control for content and dynamics when generating videos. We can generate random dynamics for fixed content, and random content for fixed dynamics. This gives us a controlled way of manipulating a video/audio sequence, such as swapping the identity of moving objects or the voice of a speaker.
\item {Efficient encoding.} Our representation is more data efficient than encoding a video frame by frame. By factoring out a separate variable that encodes content, our dynamical latent variables can have smaller dimensions. This may be promising when it comes to end-to-end neural video encoding methods.
\item We design a new metric that allow us to verify disentanglement of the latent variables, by investigating the stability of an object classifier over time.
\item We give empirical evidence, based on video data of a physics simulator, that for long sequences, a stochastic transition model generates more realistic dynamics.
\end{itemize}
%
%
The paper is structured as follows. Section \ref{sec:model} introduces the generative model and the problem setting. Section \ref{sec:related} discusses related work. Section \ref{sec:exp} presents three experiments on video and speech data. Finally, Section \ref{sec:conclusion} concludes the paper and discusses future research directions.

\section{The model}
\label{sec:model}
Let $\bm{x}_{1:T} = (\bm{x}_1, \bm{x}_2, ..., \bm{x}_T)$ denote a high dimensional sequence, such as a video with $T$ consecutive frames. Also, assume the data distribution of the training sequences is $p_{\mathcal{data}}(\bm{x}_{1:T})$. In this paper, we model the observed data with a latent variable model that separates the representation of time-invariant concepts (e.g.~object identities) from those of time-varying concepts (e.g.~pose information). 

\begin{figure*}[t]
\vspace{-0.1in}
\centering
\subfigure[generator]{
\includegraphics[width=0.3\linewidth]{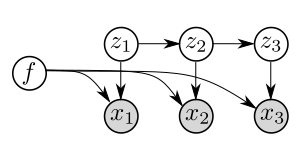}}
\subfigure[encoder (factorised $q$)]{
\includegraphics[width=0.3\linewidth]{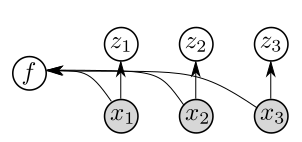}}
\subfigure[encoder (full $q$)]{
\includegraphics[width=0.3\linewidth]{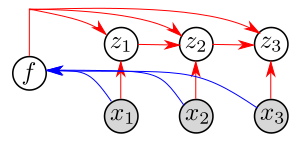}}
\caption{A graphical model visualisation of the generator and the encoder.}
\label{fig:generator}
\vspace{-0.1in}
\end{figure*}

\paragraph{Generative model.} Consider the following probabilistic model, which is also visualised in Figure \ref{fig:generator}:
\begin{equation}
\label{eq:generative_model_formula}
p_{\bm{\theta}}(\bm{x}_{1:T}, \bm{z}_{1:T}, \bm{f}) = p_{\bm{\theta}}(\bm{f})\prod_{t=1}^{T} p_{\bm{\theta}}(\bm{z}_t | \bm{z}_{<t}) p_{\bm{\theta}}(\bm{x}_t| \bm{z}_t, \bm{f}).
\end{equation}
We use the convention that $\bm{z}_0 = \bm{0}$. The generation of frame $\bm{x}_t$ at time $t$ depends on the corresponding latent variables $\bm{z}_t$ and $\bm{f}$. $\theta$ are model parameters.

Ideally, $\bm{f}$ will be capable of modelling global aspects of the whole sequence which are time-invariant, while $\bm{z}_t$ will encode time-varying features. This separation may be achieved when choosing the dimensionality of $\bm{z}_t$ to be small enough, 
thus reserving $\bm{z}_t$ only for time-dependent features while compressing everything else into $\bm{f}$. In the context of video encodings, $\bm{z}_t$ would thus encode a ``morphing transformation'', which encodes how a frame at time $t$ is morphed into a frame at time $t+1$. 

\paragraph{Inference models.} We use variational inference to learn an approximate posterior over latent variables given data~\citep{jordan1999introduction}. This involves an approximating distribution $q$. We train the generative model with the VAE algorithm~\citep{kingma2013auto}:
\begin{equation}
\max_{\bm{\theta}, \bm{\phi}} \quad \mathbb{E}_{p_{\mathcal{D}}(\bm{x}_{1:T})} \left[ \mathbb{E}_{q_{\bm{\phi}}} \left[ \log \frac{p_{\bm{\theta}}(\bm{x}_{1:T}, \bm{z}_{1:T}, \bm{f}) }{q_{\bm{\phi}}(\bm{z}_{1:T}, \bm{f} | \bm{x}_{1:T})} \right] \right].
\end{equation}
To quantify the effect of the architecture of $q$ on the learned generative model, we test with two types of $q$ factorisation structures as follows.

The first architecture constructs a factorised $q$ distribution
\begin{equation}
q_{\bm{\phi}}(\bm{z}_{1:T}, \bm{f} | \bm{x}_{1:T}) = q_{\bm{\phi}}(\bm{f} | \bm{x}_{1:T}) \prod_{t=1}^T q_{\bm{\phi}}(\bm{z}_t | \bm{x}_t)
\end{equation}
as the amortised variational distribution. We refer to this as ``factorised $q$'' in the experiments section. This factorization assumes that content features are approximately independent of motion features. Furthermore, note that the distribution over content features is conditioned on the entire time series, whereas the dynamical features are only conditioned on the individual frames.

The second encoder assumes that the variational posterior of $\z_{1:T}$ depends on $\f$, and the $q$ distribution has the following architecture:
\begin{equation}
q_{\bm{\phi}}(\bm{z}_{1:T}, \bm{f} | \bm{x}_{1:T}) = q_{\bm{\phi}}(\bm{f} | \bm{x}_{1:T}) q_{\bm{\phi}}(\bm{z}_{1:T} | \f, \bm{x}_{1:T}),
\end{equation}
and the distribution $q(\bm{z}_{1:T} | \f, \bm{x}_{1:T})$ is conditioned on the entire time series. It can be implemented by e.g.~a bi-directional LSTM \cite{graves2005framewise} conditioned on $\f$, followed by an RNN taking the bi-LSTM hidden states as the inputs. We provide a visualisation of the corresponding computation graph in the appendix. This encoder is referred to as ``full $q$''. The idea behind the structured approximation is that content may affect dynamics: in video, the shape of objects may be informative about their motion patterns, thus $\z_{1:T}$ is conditionally dependent on ${\bf f}$. The architectures of the generative model and both encoders are visualised in Figure \ref{fig:generator}.

\paragraph{Unconditional generation.}
After training, one can use the generative model to synthesise video or audio sequences by sampling the latent variables from the prior and decoding them. Furthermore, the proposed generative model allows generation of multiple sequences entailing the same global information (e.g.~the same object in a video sequence), simply by fixing $\f \sim p(\f)$, sampling different $\z^{k}_{1:T} \sim p(\z_{1:T}), k = 1, ..., K$, and generating the observations $\x^{k}_t \sim p(\x_t | \z^{k}_t, \f)$. Generating sequences with similar dynamics is done analogously, by fixing $\z_{1:T} \sim p(\z_{1:T})$ and sampling $\f^{k}, k=1, ... K$ from the prior.

\paragraph{Conditional generation.}
Together with the encoder, the model also allows conditional generation of sequences. As an example, given a video sequence $\x_{1:T}$ as reference, one can manipulate the latent variables and generate new sequences preserving either the object identity or the pose/movement information. This is done by conditioning on $\f \sim q(\f | \x_{1:T})$ for a given $\x_{1:T}$ then randomising $\z_{1:T}$ from the prior, or the other way around. 

\paragraph{Feature swapping.}

One might also want to generate a new video sequence with the object identity and pose information encoded from different sequence. Given two sequences $\x^{a}_{1:T}$ and $\x^{b}_{1:T}$, the synthesis process first infers the latent variables $\f^a \sim q(\f|\x^a_{1:T})$ and $\z^b_{1:T} \sim q(\z_{1:T}|\x^b_{1:T})$\footnote{For the full $q$ encoder it also requires $\f^b \sim q(\f|\x^b_{1:T})$.}, then produces a new sequence by sampling $\x^{\textrm{new}}_t \sim p(\x_t | \z^b_t, \f^a)$. This allows us to control both the content and the dynamics of the generated sequence, which can be applied to e.g.~conversion of voice of the speaker in a speech sequence.

\section{Related work}
\label{sec:related}
Research on learning disentangled representation has mainly focused on two aspects: the training objective and the generative model architecture. Regarding the loss function design for VAE models, \citet{higgins2016early} propose the $\beta$-VAE by scaling up the $\mathrm{KL}[q(\z|\x)|| p(\z)]$ term in the variational lower-bound with $\beta > 1$ to encourage learning of independent attributes (as the prior $p(\z)$ is usually factorised). While the $\beta$-VAE has been shown effective in learning better representations for natural images and might be able to further improve the performance of our model, we do not test this recipe here to demonstrate that disentanglement can be achieved by a careful model design.

For sequence modelling, a number of prior publications have extended VAE to video and speech data \citep{fabius2014variational, bayer2014learning, chung2015recurrent}. These models, although being able to generate realistic sequences, do not explicitly disentangle the representation of time-invariant and time-dependent information. Thus it is inconvenient for these models to perform tasks such as controlled generation and feature swapping. 

For GAN-like models, both \citet{villegas2017decomposing} and \citet{denton2017unsupervised} proposed an auto-encoder architecture for next frame prediction, with two separate encoders responsible for content and pose information at each time step. While in \citet{villegas2017decomposing}, the pose information is extracted from the difference between two consecutive frames $\x_{t-1}$ and $\x_{t}$, \citet{denton2017unsupervised} directly encoded $\x_t$ for both pose and content, and further designed a training objective to encourage learning of disentangled representations.
On the other hand, \citet{vondrick2016generating} used a spatio-temporal convolutional architecture to disentangle a video scene's foreground from its background. Although it has successfully achieved disentanglement, we note that the time-invariant information in this model is predefined to represent the background, rather than learned from the data automatically. Also this architecture is suitable for video sequences only, unlike our model which can be applied to any type of sequential data.

Very recent work \citep{hsu2017unsupervised} introduced the \emph{factorised hierarchical variational auto-encoder} (FHVAE) for unsupervised learning of disentangled representation of speech data. Given a speech sequence that has been partitioned into segments $\{ \x_{1:T}^{n}\}_{n=1}^N$, FHVAE models the joint distribution of $\{ \x_{1:T}^{n}\}_{n=1}^N$ and latent variables as follows:
\begin{equation*}
\begin{aligned}
p(\{ \x_{1:T}^{n}, \z_1^{n}, \z_2^{n} \}, \bm{\mu}_2) = p(\bm{\mu}_2) \prod_{n=1}^N p(\x^{n}_{1:T}, \z^{n}_1, \z^{n}_2| \bm{\mu}_2), \\
p(\x^{n}_{1:T}, \z^{n}_1, \z^{n}_2| \bm{\mu}_2) =  p(\z^{n}_1) p(\z^{n}_2 | \bm{\mu}_2) p(\x^{n}_{1:T} | \z^{n}_1, \z^{n}_2).
\end{aligned}
\end{equation*}
Here the $\z^{n}_2$ variable has a hierarchical prior $p(\z^{n}_2 | \bm{\mu}_2) = \mathcal{N}(\bm{\mu}_2, \sigma^2 \mathbf{I})$, $p(\bm{\mu}_2) = \mathcal{N}(\bm{0}, \lambda \mathbf{I})$. The authors showed that by having different prior structures for $\z^{n}_1$ and $\z^{n}_2$, it allows the model to encode with $\z^{n}_2$ speech sequence-level attributes (e.g.~pitch of a speaker), and other residual information with $\z^{n}_1$. A discriminative training objective (see discussions in Section \ref{sec:exp-speech}) is added to the variational lower-bound, which has been shown to further improve the quality of the disentangled representation. Our model can also benefit from the usage of hierarchical prior distributions, e.g.~$\f^{n} \sim p(\f | \bm{\mu}_2), \bm{\mu}_2 \sim p(\bm{\mu}_2)$, and we leave the investigation to future work.

\section{Experiments}
\label{sec:exp}
We carried out experiments both on video data (Section~\ref{sec:exp-video}) as well as speech data (Section~\ref{sec:exp-speech}). In both setups, we find strong evidence that our model learns an approximately disentangled representation that allows for conditional generation and feature swapping. We further investigated the efficiency for encoding long sequences with a stochastic transition model in Section \ref{sec:exp-ball}. The detailed model architectures of the networks used in each experiment are reported in the appendix.
\subsection{Video sequence: Sprites}
\label{sec:exp-video}
We present an initial test of the proposed VAE architecture on a dataset of video game ``sprites'', i.e. animated cartoon characters whose clothing, pose, hairstyle, and skin color we can fully control. This dataset comes from an open-source video game project called Liberated Pixel Cup\footnote{\url{http://lpc.opengameart.org/}}, and has been also considered in \citet{reed2015deep,mathieu2016disentangling} for image processing experiments. 
Our experiments show that static attributes such as hair color and clothing are well preserved over time for randomly generated videos. 

\begin{figure}[t]
\centering
\includegraphics[width=1.0\linewidth]{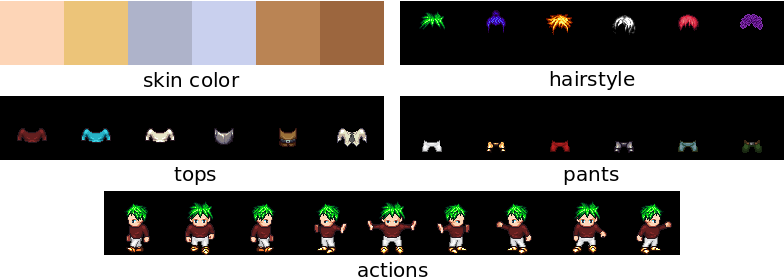}
\vspace{-0.1in}
\caption{A visualisation of the attributes and actions used to generate the Sprite data set. See main text for details.}
\label{fig:attributes}
\vspace{-0.1in}
\end{figure}

\paragraph{Data and preprocessing.} We downloaded and selected the online available sprite sheets\footnote{\url{https://github.com/jrconway3/Universal-LPC-spritesheet}}, and organised them into 4 attribute categories (skin color, tops, pants and hairstyle) and 9 action categories (walking, casting spells and slashing, each with three viewing angles). In order to avoid a combinatorial explosion problem, each of the attribute categories contains 6 possible variants (see Figure \ref{fig:attributes}), therefore it leads to $6^4 = 1296$ unique characters in total. We used $1000$ of them for training/validation and the rest of them for testing. The resulting dataset consists of sequences with $T=8$ frames of dimension $64 \times 64$. Note here we did not use the labels for training the generative model. Instead these labels on the data frames are used to train a classifier that is later deployed to produce quantitative evaluations on the VAE, see below.

\begin{figure}[t]
\centering
\vspace{-0.1in}
\subfigure[random test data sequences]{
\includegraphics[width=0.475\linewidth]{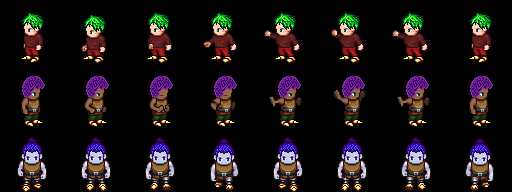}}
\subfigure[reconstruction]{
\includegraphics[width=0.475\linewidth]{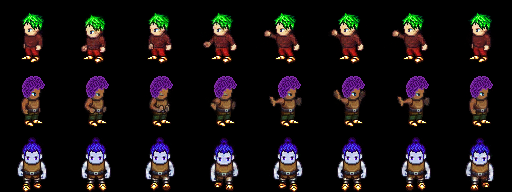}
}
\vspace{-0.1in}
\subfigure[reconstruction with randomly sampled $\f$]{
\includegraphics[width=0.475\linewidth]{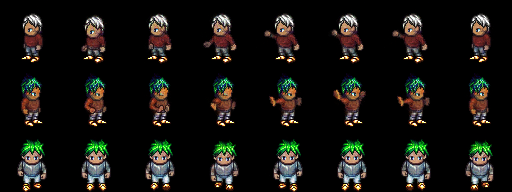}}
\subfigure[reconstruction with randomly sampled $\z_{1:T}$]{
\includegraphics[width=0.475\linewidth]{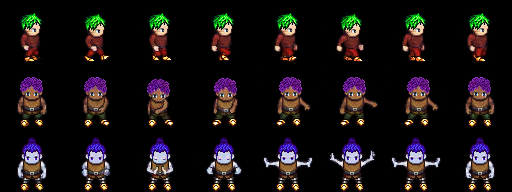}}
\vspace{-0.1in}
\subfigure[reconstruction with swapped encoding $\f$]{
\includegraphics[width=0.475\linewidth]{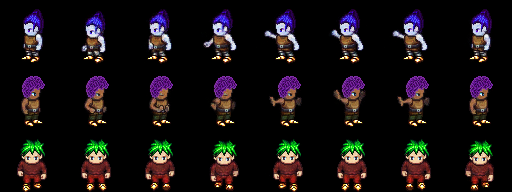}}
\subfigure[reconstruction with swapped encoding $\z_{1:T}$]{
\includegraphics[width=0.475\linewidth]{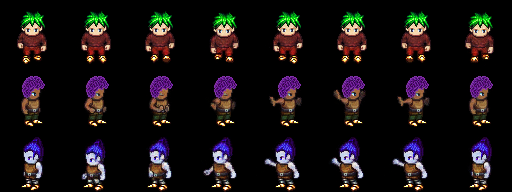}}
\subfigure[generated sequences with fixed $\f$]{
\includegraphics[width=0.475\linewidth]{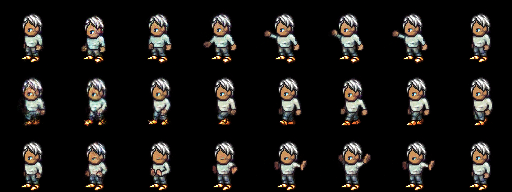}}
\subfigure[generated sequences with fixed $\z_{1:T}$]{
\includegraphics[width=0.475\linewidth]{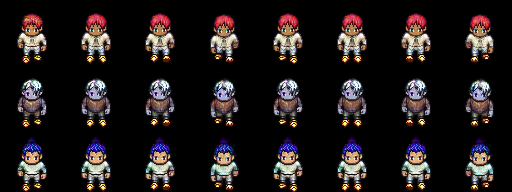}}
\caption{Visualisation of generated and reconstructed video sequences. See main text for discussions.}
\label{fig:visualisation}
\vspace{-0.1in}
\end{figure}

\paragraph{Qualitative analysis.} We start with a qualitative evaluation of our VAE architecture. Figure 
\ref{fig:visualisation} shows both reconstructed as well as generated video sequences from our model. Each panel shows three video sequences with time running from left to right. Panel (a) shows parts of the original data from the test set, and (b) shows its reconstruction. 

The sequences visualised in panel (c) are generated using $\bm{z}_t \sim q(\bm{z}_t |\bm{x}_t)$ but $\bm{f} \sim p(\bm{f})$. Hence, the dynamics are imposed by the encoder, but the identity is sampled from the prior. We see that panel (c) reveals the same motion patterns as (a), but has different character identities. Conversely, in panel (d) we take the identity from the encoder, but sample the dynamics from the prior. Panel (d) reveals the same characters as (a), but different motion patterns.

Panels (e) and (f) focus on feature swapping. In (e), the frames are constructed by computing $\bm{z}_t \sim q(\bm{z}_t | \bm{x}_t)$ on one input sequence but $\bm{f}$ encoded on another input sequence. These panels demonstrate that the encoder and the decoder have learned a factored representation for content and pose. 

Panels (g) and (h) focus on conditional generation, showing randomly generated sequences that share the same $\bm{f}$ or $\bm{z}_{1:T}$ samples from the prior. Thus, in panel (g) we see the same character performing different actions, and in (h) different characters performing the same motion. This again illustrates that the prior model disentangles the representation.

\paragraph{Quantitative analysis.} Next we perform quantitative evaluations of the generative model, using a classifier trained on the labelled frames. 
Empirically, we find that the fully factorized and structured inference networks produce almost identical results here, presumably because in this dataset the object identity and pose information are truly independent. Therefore we only report results on the fully factorised $q$ distribution case.

The first evaluation task considers reconstructing the test sequences with encoded $\bm{f}$ and randomly sampled $\bm{z}_t$ (in the same way as to produce panel (d) in Figure \ref{fig:visualisation}). Then we compare the classifier outputs on both the original frames and the reconstructed frames. If the character's identity is preserved over time, the classifier should produce identical probability vectors on the data frames and the reconstructed frames (denoted as $\bm{p}_{data}$ and $\bm{p}_{recon}$ respectively).

We evaluate the similarity between the original and reconstructed sequences both in terms of the disagreement of the predicted class labels $\max_{i}[\bm{p}_{recon}(i)] \neq \max_{i}[\bm{p}_{data}(i)]$ and the KL-divergence $\mathrm{KL}[\bm{p}_{recon} || \bm{p}_{data}]$.
We also compute the two metrics on the action predictions using reconstructed sequences with randomised $\bm{f}$ and inferred $\bm{z}_t$. The results in Table \ref{tab:recon_test} indicate that the learned representation is indeed factorised. For example, in the fix-$\bm{f}$ generation test, only $4\%$ out of $296 \times 9$ data-reconstruction frame pairs contain characters whose generated skin color differs from the rest, where in the case of hairstyle preservation the disagreement rate is only $0.06\%$. The KL metric is also much smaller than the KL-divergence $\mathrm{KL}[\bm{p}_{random} || \bm{p}_{data}]$ where $\bm{p}_{random} = (1/N_{\text{class}}, ..., 1/N_{\text{class}})$, indicating that our result is significant.

\begin{figure}
\centering
\vspace{-0.1in}
\subfigure[\label{fig:gen_fix_f} Trajectory plots on the generated sequences with shared $\bm{f}$.]{
\includegraphics[width=1.0\linewidth]{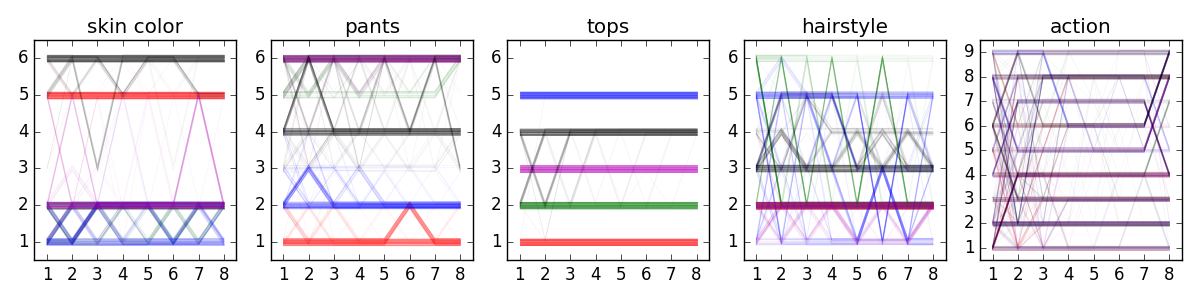}}
\vspace{-0.1in}
\subfigure[\label{fig:gen_fix_z} Trajectory plots on the generated sequences with shared $\bm{z}_{1:T}$.]{
\includegraphics[width=1.0\linewidth]{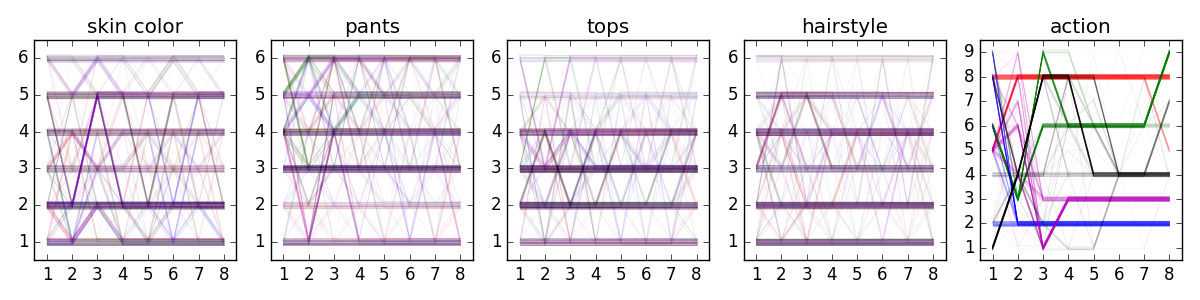}}
\caption{Classification test on the generated video sequences with shared $\bm{f}$ (top) or shared $\bm{z}_{1:T}$ (bottom), respectively. The experiment is repeated 5 times and depicted by different color coding. The x and y axes are time and the class id of the attributes, respectively.}
\end{figure}

In the second evaluation, we test whether static attributes of generated sequences, such as clothing or hair style, are preserved over time. We sample 200 video sequences from the generator, using the same $\bm{f}$ but different latent dynamics $\bm{z}_{1:T}$. We use the trained classifier to predict both the attributes and the action classes for each of the generated frames. Results are shown in Figure \ref{fig:gen_fix_f}, where we plot the prediction of the classifiers for each frame over time. For example, the trajectory curve in the ``skin color'' panel in Figure \ref{fig:gen_fix_f} corresponds to the skin color attribute classification results for frames $\x_{1:T}$ of a generated video sequence. We repeat this process 5 times with different $\f$ samples, where each $\f$ corresponds to one color.


It becomes evident that those lines with the same color are clustered together, confirming that $\f$ mainly controls the generation of time-invariant attributes.
Also, most character attributes are preserved over time, e.g. for the attribute ``tops'', the trajectories are mostly straight lines. 
However, some of the trajectories for the attributes drift away from the majority class. We conjecture that this is due of the mass-covering behaviour of (approximate) maximum likelihood training, which makes the trained model generate characters that do not exist in the dataset. Indeed the middle row of panel (c) in Figure \ref{fig:visualisation} contains a character with an unseen hairstyle, showing that our model is able to generalise beyond the training set. 
On the other hand, the sampling process returns sequences with diverse actions as depicted in the action panel, meaning that $\f$ contains little information regarding the video dynamics. 

\begin{figure}
\centering
\includegraphics[width=0.6\linewidth]{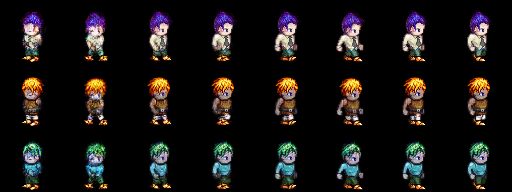}
\caption{Visualising multi-modality in action space. In this case the characters turn from left to right, and this action sequence is not observed in data.}
\label{fig:gen_fix_z_multi}
\vspace{-0.1in}
\end{figure}

We performed similar tests on sequence generations with shared latent dynamics $\bm{z}_{1:T}$ but different $\bm{f}$, shown in Figure \ref{fig:gen_fix_z}. 
The experiment is repeated 5 times as well, and again trajectories with the same color encoding correspond to videos generated with the same $\z_{1:T}$ (but different $\f$). 
Here we also observe diverse trajectories for the attribute categories. In contrast, the characters' actions are mostly the same. These two test results again indicate that the model has successfully learned disentangled representations of character identities and actions.
Interestingly we observe multi-modalities in the action domain for the generated sequences, e.g. the trajectories in the action panel of Figure \ref{fig:gen_fix_z} are jumping between different levels. We also visualise in Figure \ref{fig:gen_fix_z_multi} generated sequences of the ``turning'' action that is not present in the dataset. It again shows that the trained model generalises to unseen cases.

\begin{table}[t]
\centering
\vspace{-0.1in}
\caption{Averaged classification disagreement and KL similarity measures for our model on Sprite data. Note here KL-recon = $\mathrm{KL}[\bm{p}_{recon} || \bm{p}_{data}]$ and KL-random = $\mathrm{KL}[\bm{p}_{random} || \bm{p}_{data}]$. \vspace{0.1in}}
\label{tab:recon_test}
\begin{tabular}{l|ccc}
\hline
attributes & disagreement & KL-recon & KL-random  \\
\hline
skin colour & 3.98\% & 0.7847 & 8.8859 \\
pants & 1.82\% & 0.3565 & 8.9293 \\
tops & 0.34\% & 0.0647 & 8.9173 \\
hairstyle & 0.06\% & 0.0126 & 8.9566 \\
\hline
action & 8.11\% & 0.9027 & 13.7510 \\
\hline
\end{tabular}
\vspace{-0.1in}
\end{table}

\subsection{Speech data: TIMIT}
We also experiment on audio sequence data. Our disentangled representation allows us to convert speaker identities into each other while conditioning on the content of the speech. We also show that our model gives rise to speaker verification, where we outperform a recent probabilistic baseline model.

\label{sec:exp-speech}
\paragraph{Data and preprocessing.}
The TIMIT data \cite{garofolo1993timit} contains broadband
16kHz recordings of phonetically-balanced read speech. A total of 6300 utterances (5.4 hours) are presented with 10 sentences from each of the 630 speakers (70\% male and
30\% female).
We follow \citet{hsu2017unsupervised} for data pre-processing: the raw speech waveforms are first split into sub-sequences of 200ms, and then preprocessed with sparse fast Fourier transform to obtain a 200 dimensional log-magnitude spectrum, computed every 10ms. This implies $T=20$ for the observation $\x_{1:T}$.

\begin{figure}[t]
\centering
\subfigure[female speech (original)]{
\includegraphics[width=0.475\linewidth]{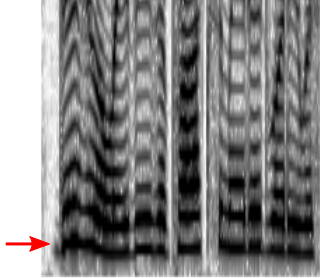}}
\subfigure[female to male]{
\includegraphics[width=0.475\linewidth]{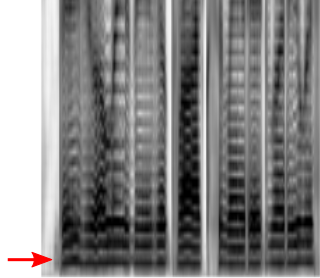}}
\subfigure[male speech (original)]{
\includegraphics[width=0.475\linewidth]{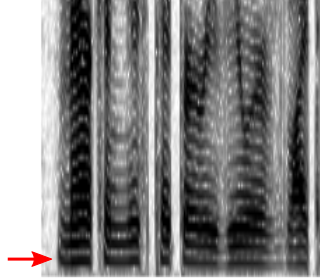}
}
\subfigure[male to female]{
\includegraphics[width=0.475\linewidth]{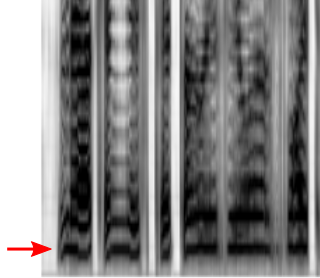}}
\caption{Visualising the spectrum of the reconstructed speech sequences. Here we show the spectrogram for the first 2000ms, with horizontal axis denoting time and the vertical axis denoting frequencies. The red arrow points to the first harmonics which indicates the fundamental frequency of the speech signal.}
\label{fig:timit_visualisation}
\vspace{-0.1in}
\end{figure}

\paragraph{Qualitative analysis.}
We perform voice conversion experiments to demonstrate the disentanglement of the learned representation. The goal here is to convert male voice to female voice (and vice versa) with the speech content being preserved. Assuming that $\f$ has learned the representation of speaker's identity, the conversion can be done by first encoding two sequences $\x^{\textrm{male}}_{1:T}$ and $\x^{\textrm{female}}_{1:T}$ with $q$ to obtain representations $\{\f^{\textrm{male}}, \z^{\textrm{male}}_{1:T} \}$ and $\{\f^{\textrm{female}}, \z^{\textrm{female}}_{1:T} \}$, then construct the converted sequence by feeding $\f^{\textrm{female}}$ and $\z^{\textrm{male}}_{1:T}$ to the decoder $p(\x_t|\z_t, \f)$. Figure \ref{fig:timit_visualisation} shows the reconstructed spectrogram after the swapping process of the $\f$ features. We also provide the reconstructed speech waveforms using the Griffin-Lim algorithm \cite{griffin1984signal} in the appendix. 

The experiments show that the harmonics of the converted speech sequences shifted to higher frequency in the ``male to female'' test and vice versa. Also the pitch (the red arrow in Figure~\ref{fig:timit_visualisation} indicating the fundamental frequency, i.e.~the first harmonic) of the converted sequence (b) is close to the pitch of (c), same as for the comparison between (d) and (a). By an informal listening test of the speech sequence pairs (a, d) and (b, c), we confirm that the speech content is preserved. 
These results show that our model is successfully applied to speech sequences for learning disentangled representations.

\paragraph{Quantitative analysis.}
We further follow \citet{hsu2017unsupervised} to use speaker verification for quantitative evaluation. Speaker verification is the process of verifying the claimed
identity of a speaker, usually by comparing the ``features'' $\bm{w}^{\textrm{test}}$ of the test utterance $\x^{\textrm{test}}_{1:T_1}$ with those of the target utterance $\x^{\textrm{target}}_{1:T_2}$ from the claimed identity. The claimed identity is confirmed if the cosine similarity  $\textrm{cos}(\bm{w}^{\textrm{test}}, \bm{w}^{\textrm{target}})$ is grater than a given threshold $\epsilon$ \citep{dehak2009support}. By varying $\epsilon \in [0, 1]$, we report the verification performance in terms of equal error rate (EER), where the false rejection rate equals the false acceptance rate. 

The extraction of the ``features'' is crucial for the performance of this speaker verification system. Given a speech sequence containing $N$ segments $\{ \x^{(n)}_{1:T} \}_{n=1}^N$, we constructed two types of ``features'', one by computing $\bm{\mu}_{\f}$ as the mean of $q(\f^{(n)} | \x^{(n)}_{1:T})$ across the segments, and the other by extracting the mean $\bm{\mu}_{\z_t}$ of $q(\z_t| \x_{1:T})$ and averaging them across both time $T$ and segments. In formulas, 
$$
\bm{\mu}_{\f} = \frac{1}{N} \sum_{n=1}^N \bm{\mu}_{\f^{n}}, \quad \bm{\mu}_{\f^{n}} = \mathbb{E}_{q(\f^{n} | \x^{n}_{1:T}) }[\f^{n}],
$$
$$
\bm{\mu}_{\z} = \frac{1}{TN} \sum_{t=1}^T \sum_{n=1}^N \bm{\mu}_{\z^{n}_t}, \quad 
\bm{\mu}_{\z^{n}_t} = \mathbb{E}_{q(\z^{n}_t | \x^{n}_{1:T})}[\z^{n}_t].
$$
We also include two baseline results from \citet{hsu2017unsupervised}: one used the i-vector method \citep{dehak2011front} for feature extraction, and the other one used $\bm{\mu}_1$ and $\bm{\mu}_2$ (analogous to $\bm{\mu}_{\z}$ and $\bm{\mu}_{\f}$ in our case) from a trained FHVAE model on Mel-scale filter bank (FBank) features.

The test data were created from the test set of TIMIT, containing 24 unique speakers and 18,336 pairs for verification. Table \ref{tab:timit_verify} presents the EER results of the proposed model and baselines.\footnote{ \citet{hsu2017unsupervised} did not 	provide the EER results for $\alpha = 0$ and $\bm{\mu}_1$ in the 16 dimension case.}
It is clear that the $\bm{\mu}_{\f}$ feature performs significantly better than the i-vector method,  indicating that the $\f$ variable has learned to represent a speaker's identity. On the other hand, using $\bm{\mu}_{\z}$ as the features returns considerably worse EER rates compared to the i-vector method and $\bm{\mu}_{\f}$ feature. This is good, as it indicates that the $\z$ variables contain less information about the speaker's identity, again validating the success of disentangling time-variant and time-independent information. Note that the EER results for $\bm{\mu}_{\z}$ get worse when using the full $q$ encoder, and in the 64 dimensional feature case the verification performance of $\bm{\mu}_{\f}$ improves slightly. This also shows that for real-world data it is useful to use a structured inference network to further improve the quality of disentangled representation. 

Our results are competitive with (or slightly better than) the FHVAE results ($\alpha=0$) reported in \citet{hsu2017unsupervised}. The better results for FHVAE ($\alpha=10$) is obtained by adding a discriminative training objective (scaled by $\alpha$) to the variational lower-bound. 
In a nutshell, the time-invariant information in FHVAE is encoded in a latent variable $\z^n_{2} \sim p(\z^n_2|\bm{\mu}_2)$, and the discriminative objective encourages $\z^n_2$ encoded from a segment of one sequence to be close to the corresponding $\bm{\mu}_2$ while far away from $\bm{\mu}_2$ of other sequences. 
However, we do not test this idea here because (1) our goal is to demonstrate that the proposed architecture is a minimalistic framework for learning disentangled representations of sequential data; (2) this discriminative objective is specifically designed for hierarchical VAE, and in general the assumption behind it might not always be true (consider encoding two speech sequences coming from the same speaker). 
Similar ideas for discriminative training have been considered in e.g.~\citet{mathieu2016disentangling}, but that discriminative objective can only be applied to two sequences that are known to entail different time-invariant information (e.g.~two sequences with different labels), which implicitly uses supervisions. Nevertheless, a better design for the discriminative objective without supervision can further improve the disentanglement of the learned representations, and we leave it to future work.

\begin{table}
\centering
\vspace{-0.1in}
\caption{Speaker verification errors, comparing the FHVAE with our approach. Static information is encoded in $\bm{\mu}_2$ / $\bm{\mu}_{\f}$ and dynamic information in $\bm{\mu}_1$ / $\bm{\mu}_{\z}$ for the FHVAE / our approach, respectively. Large errors are expected when predicting based on $\bm{\mu}_1$ / $\bm{\mu}_{\z}$, and small errors for $\bm{\mu}_2$ / $\bm{\mu}_{\f}$, respectively (see main text).
Our data-agnostic approach compares favourably.
}
\label{tab:timit_verify}
\begin{tabular}{lccc}
\hline
model & feature & dim & EER  \\
\hline
- & i-vector & 200 & 9.82\% \\
\hline
FHVAE ($\alpha=0$) & $\bm{\mu}_2$ & 16 & 5.06\% \\
FHVAE ($\alpha=10$) & $\bm{\mu}_2$ & 32 & 2.38\% \\
                    & $\bm{\mu}_1$ & 32 & 22.47\% \\
\hline
factorised q & $\bm{\mu}_{\f}$ & 16 & 4.78\% \\
         & $\bm{\mu}_{\z}$ & 16 & 17.84\% \\
factorised q & $\bm{\mu}_{\f}$ & 64 & 4.94\% \\
         & $\bm{\mu}_{\z}$ & 64 & 17.49\% \\
full q & $\bm{\mu}_{\f}$ & 16 & 5.64\% \\
         & $\bm{\mu}_{\z}$ & 16 & 19.20\% \\
full q & $\bm{\mu}_{\f}$ & 64 & 4.82\% \\
       & $\bm{\mu}_{\z}$ & 64 & 18.89\% \\
\hline
\end{tabular}
\end{table}

\subsection{Comparing stochastic \& deterministic dynamics}
\label{sec:exp-ball}

Lastly, although not a main focus of the paper, we show that the usage of a \emph{stochastic} transition model for the prior leads to more realistic dynamics of the generated sequence. For comparison, we consider another class of models:
\begin{equation*}
\label{eq:baseline_recurrence}
p(\x_{1:T}, \z, \f) = p(\f)p(\z) \prod_{t=1}^T p(\x_t|\z, \f). 
\end{equation*}
The parameters of $p(\x_t | \z, \f)$ are defined by a neural network $\text{NN}(\bm{h}_t, \f)$, with $\bm{h}_t$ computed by a \emph{deterministic} RNN conditioned on $\z$. 
We experiment with two types of deterministic dynamics. The first model uses an LSTM with $\z$ as the initial state: $\bm{h}_0 = \z$, $\bm{h}_t = \text{LSTM}(\bm{h}_{t-1})$. In later experiments we refer this dynamics as \textbf{LSTM-f} as the latent variable $\z$ is forward propagated in a deterministic way. The second one deploys an LSTM conditioned on $\z$ (i.e. $\bm{h}_0 = \bm{0}, \bm{h}_t = \text{LSTM}(\bm{h}_{t-1}, \z)$), therefore we refer it as \textbf{LSTM-c}. This is identical to the transition dynamics used in the FHVAE model \cite{hsu2017unsupervised}. 
For comparison, we refer to our model as the 'stochastic' model (Eq.~\ref{eq:generative_model_formula}).

The LSTM models encodes temporal information in a global latent variable $\z$. Therefore, small differences/errors in $\bm{z}$ will accumulate over time, which may result in unrealistic long-time dynamics. In contrast, the stochastic model (Eq.~\ref{eq:generative_model_formula}) keeps track of the time-varying aspects of $\x_t$ in $\z_t$ for every $t$, making the reconstruction to be time-local and therefore much easier. Therefore, the stochastic model is better suited if the sequences are long and complex. We give empirical evidence to support this claim.

\paragraph{Data preprocessing \& hyper-parameters.}
We follow \citet{fraccaro2017disentangled} to simulate video sequences of a ball (or a square) bouncing inside an irregular polygon using Pymunk.\footnote{\url{http://www.pymunk.org/en/latest/}. For simplicity we disabled rotation of the square when hitting the wall, by setting the inertia to infinity.} 
The irregular shape was chosen because it induces chaotic dynamics, meaning that small deviations from the initial position and velocity of the ball will create exponentially diverging trajectories at long times. This makes memorizing the dynamics of a prototypical sequence challenging. 
%
We randomly sampled the initial
position and velocity of the ball, but did not apply any force to the ball, except for the fully elastic collisions with the walls. We generated 5,000 sequences in total (1000 for test), each of them containing $T=30$ frames with a resolution of $32 \times 32$.
For the deterministic LSTMs, we fix the dimensionality of $\z_t$ to 64, and set $\bm{h}_t$ and the LSTM internal states to be 512 dimensions. The latent variable dimensionality of the stochastic dynamics is $\text{dim}(\z_t) = 16$.

\begin{figure}[t]
\centering
\subfigure[data for reconstruction]{
\includegraphics[width=0.475\linewidth]{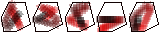}}
\subfigure[data for prediction]{
\includegraphics[width=0.475\linewidth]{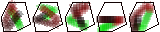}
}
\subfigure[reconstruction (stochastic)]{
\includegraphics[width=0.475\linewidth]{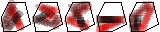}}
\subfigure[prediction (stochastic)]{
\includegraphics[width=0.475\linewidth]{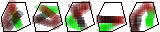}}
\subfigure[reconstruction (LSTM-f)]{
\includegraphics[width=0.475\linewidth]{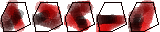}}
\subfigure[prediction (LSTM-f)]{
\includegraphics[width=0.475\linewidth]{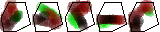}}
\subfigure[reconstruction (LSTM-c)]{
\includegraphics[width=0.475\linewidth]{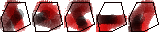}}
\subfigure[prediction (LSTM-c)]{
\includegraphics[width=0.475\linewidth]{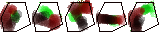}}
\caption{Predicted and reconstructed video sequences. The videos are shown as single images, with colour intensity (starting from black) representing the incremental sequence index ('stochastic' is proposed). The missing/predicted frames are shown in green.}
\vspace{-0.1in}
\label{fig:ball_visualisation}
\end{figure}

\begin{figure}
\includegraphics[width=0.95\linewidth]{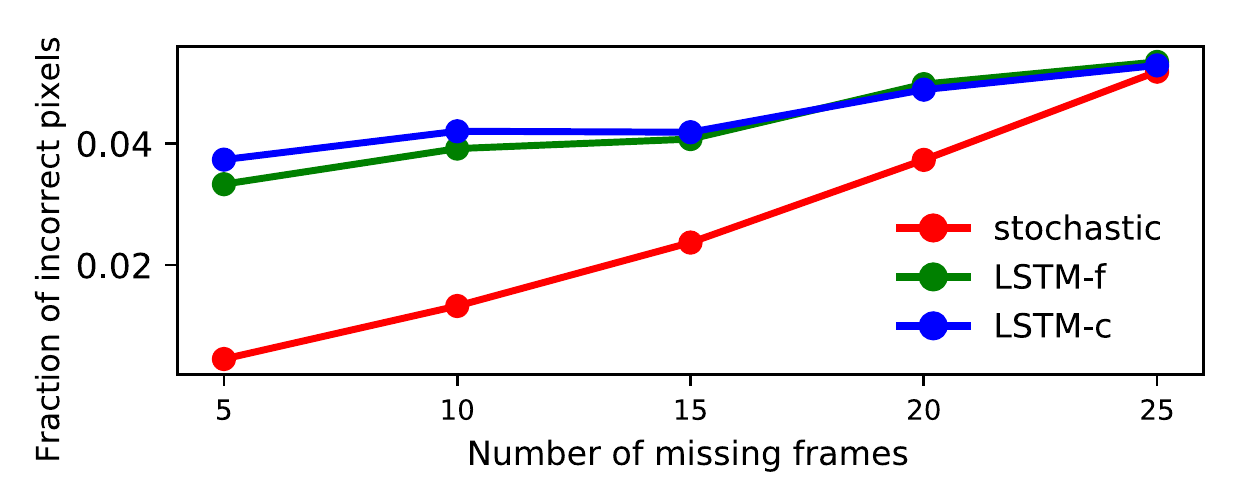}
\vspace{-0.1in}
\caption{Reconstruction error of different models as a function of the number of consecutive missing frames (see main text). 
Lower values are better. 'stochastic' refers to the proposed approach.}
\label{fig:ball_quantitative}
\vspace{-0.1in}
\end{figure}

\paragraph{Qualitative \& quantitative analyses.} 
We consider both reconstruction and missing data imputation tasks for the learned generative models. For the latter and for $T=30$, the models observe the first $t < T$ frames of a sequence and predict the remaining $T - t$ frames using the prior dynamics. We visualise in Figure \ref{fig:ball_visualisation} the ground truth, reconstructed, and predicted sequences ($t=20$) from all models. We further consider average fraction of incorrectly reconstructed/predicted pixels as a quantitative metric, to evaluate how well the ground-truth dynamics is recovered given consecutive missing frames. The result is reported in Figure \ref{fig:ball_quantitative}.
The stochastic model outperforms the deterministic models both qualitatively and quantitatively. The shape of the ball is better preserved over time, and the trajectories are more physical. This explains the lower errors of the stochastic model, and the advantage is significant when the number of missing frames is small.

Our experiments give evidence that the stochastic model is better suited to modelling long, complex sequences when compared to the deterministic dynamical models. We expect that a better design for the stochastic transition dynamics, e.g.~by combining deep neural networks with well-studied linear dynamical systems \cite{krishnan2015deep, fraccaro2016sequential, karl2016deep,johnson2016composing,krishnan2017structured,fraccaro2017disentangled}, can further enhance the quality of the learned representations.


%

\section{Conclusions and outlook}
\label{sec:conclusion}

We presented a minimalistic generative model for learning disentangled representations of high-dimensional time series. Our  model consists of a global  latent variable for content features, and a stochastic RNN with time-local latent variables for dynamical features. The model is trained using standard amortized variational inference.
We carried out experiments both on video and audio data. Our approach allows us to perform full and conditional generation, as well as feature swapping, such as voice conversion and video content manipulation.  We also showed that a stochastic transition model generally outperforms a deterministic one.

Future work may investigate whether a similar model applies to more complex video and audio sequences. Also, disentangling may further be improved by additional cross-entropy terms, or discriminative training. 
A promising avenue of research is to explore the usage of this architecture for neural compression. An advantage of the model is that it separates dynamical from static features, allowing the latent space for the dynamical part to be low-dimensional. 

\section*{Acknowledgements}
We thank Robert Bamler, Rich Turner, Jeremy Wong and Yu Wang for discussions and feedback on the manuscript. We also thank Wei-Ning Hsu for helping reproduce the FHVAE experiments. Yingzhen Li thanks Schlumberger Foundation FFTF fellowship for supporting her PhD study.

\bibliography{references}
\bibliographystyle{icml2018}

\onecolumn
\appendix

\section{Computation graph for the full $q$ inference network}

\begin{figure}
\vspace{-0.1in}
\centering
\subfigure[encoder for $\f$ (full $q$)]{
\includegraphics[width=0.3\linewidth]{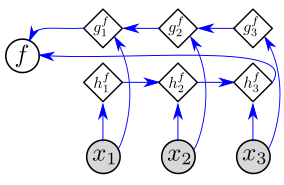}}
\hspace{1.0in}
\subfigure[encoder for $\z$ (full $q$)]{
\includegraphics[width=0.3\linewidth]{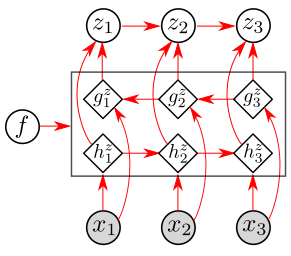}}
\caption{A graphical model visualisation of the generator and the encoder.}
\label{fig:encoder_full}
\vspace{-0.1in}
\end{figure}

In Figure \ref{fig:encoder_full} we show the computation graph of the full $q$ inference framework. The inference model first computes the mean and variance parameters of $q(\f | \x_{1:T})$ with a bi-directional LSTM \citep{graves2005framewise} and samples $\f$ from the corresponding Gaussian distribution (see Figure (a)). Then $\f$ and $\x_{1:T}$ are fed into another bi-directional LSTM to compute the hidden state representations $\bm{h}_t^z$ and $\bm{g}_t^z$ for the $\z_t$ variables (see Figure (b)), where at each time-step both LSTMs take $[\x_t, \f]$ as the input and update their hidden and internal states. Finally the parameters of $q(\z_{1:T}|\x_{1:T}, \f)$ is computed by a simple RNN with input $[\bm{h}_t^z, \bm{g}_t^z]$ at time $t$.

\section{Sound files for the speech conversion test}
We provide sound files to demonstrate the conversion of female/male speech sequences at \url{https://drive.google.com/file/d/1zpiZJNjGWw9pGPYVxgSeoipiZdeqHatY/view?usp=sharing}. Given a spectrum (magnitude information), the sound waveform is reconstructed using the Griffin-Lim algorithm \cite{griffin1984signal}, which initialises the phase randomly, then iteratively refine the phase information by looping the SFFT/inverse SFFT transformation until convergence or reaching some stopping criterion. We note that the sound quality can be further improved by e.g.~conjugate gradient methods. Also we found in general it is more challenging to convert female speech to male speech than the other way around, which is also observed by \cite{hsu2017unsupervised}.

We also note here that the phase information is not modelled in our experiments, nor in the FHVAE tests. First, as phase is a circular variable ranging from $[-\pi, \pi]$, Gaussian distribution is inappropriate, and instead a von Mises distribution is required. However, fast computation of the normalising constant of a von Mises distribution -- which is a Bessel function -- remains a challenging task, let alone differentiation and optimisation of the concentration parameters.

\section{Network architecture}
\paragraph{Sprite.} 
The prior dynamics $p_{\bm{\theta}}(\bm{z}_t| \bm{z}_{<t})$ is Gaussian with parameters computed by an LSTM \citep{hochreiter1997long}. Then $\bm{x}_t$ is generated by a deconvolutional neural network, which first transforms $[\bm{z}_t,\bm{f}]$ with a one hidden-layer MLP, then applies 4 deconvolutional layers with 256 channels and up-sampling. We use the $\ell_2$ loss for the likelihood term, i.e.~$\log p(\bm{x}_t | \bm{z}_t, \bm{f}) = - \frac{1}{2}|| \bm{x}_t - \mathrm{NN}_{\bm{\theta}}(\bm{z}_t, \bm{f}) ||_2^2 + const$. 

For the inference model, we first use a convolutional neural network, with a symmetric architecture to the deconvolutional one, to extract visual features. Then $q(\bm{f} | \bm{x}_{1:T})$ is also a Gaussian distribution parametrised by an LSTM and depends on the entire sequence of these visual features. For the factorised $q$ encoder, $q(\bm{z}_t | \bm{x}_t)$ is also Gaussian parametrised by a one-hidden layer MLP taking the visual feature of $\x_t$ as input.
The dimensions of $\bm{f}$ and $\bm{z}_t$ are 256 and 32, respectively, and the hidden layer sizes are fixed to 512.

\paragraph{TIMIT.} 
We use almost identical architecture as in the Sprite data experiment, except that the likelihood term $p(\x_t | \z_t, \f)$ is defined as Gaussian with mean and variance determined by a 2-hidden-layer MLP taking both $\z_t$ and $\f$ as inputs. The dimensions of $\f$ and $\z_t$ are 64 if not specifically stated, and the hidden layer sizes are fixed to 256. 

For the full $q$ inference model we use the architecture visualised in Figure \ref{fig:encoder_full}. Again the bi-LSTM networks take the features of $\x_t$ as inputs, where those features are extracted using a one hidden-layer MLP.

\paragraph{Bouncing ball.}
We use an RNN (instead of an LSTM as in previous experiments) to parametrise the stochastic prior dynamics of our model, and set the dimensionality of $\z_t$ to be 16. For the deterministic models we set $\z$ to be 64 dimensional. We use a 64 dimensional $\f$ variable and Bernoulli likelihood for all models.

For inference models, we use the full $q$ model for the stochastic dynamics case. For the generative models with deterministic dynamics, we also use bi-LSTMs of the same architecture to infer the parameters of $q(\f|\x_{1:T})$ and $q(\z|\x_{1:T})$. Again a convolutional neural network is deployed to compute visual features for LSTM inputs.

All models share the same architecture of the (de-)convolution network components.
The deconvolution neural network has 3 deconvolutional layers with 64 channels and up-sampling. The convolutional neural network for the encoder has a symmetric architecture to the deconvolution one. The hidden layer sizes in all networks are fixed to 512.

\end{document}